\documentclass[10pt,twocolumn,letterpaper]{article}

\usepackage[T1]{fontenc}
\usepackage{times}
\usepackage[margin=0.72in,columnsep=0.27in]{geometry}
\usepackage{graphicx}
\usepackage{amsmath,amssymb}
\usepackage{booktabs,array,multirow}
\usepackage{xcolor}
\usepackage{microtype}
\usepackage[font=small,labelfont=bf,labelsep=period]{caption}
\usepackage[numbers,sort&compress]{natbib}
\usepackage{hyperref}

\definecolor{linkblue}{HTML}{245A80}
\hypersetup{colorlinks=true,allcolors=linkblue}

\newcommand{\method}{TL-VLA}
\newcommand{\E}{\mathbb{E}}

\newcommand{\F}{\mathbf{F}}
\newcommand{\G}{\mathbf{G}}
\newcommand{\U}{\mathbf{U}}

\title{\bfseries Beyond Natural-Language Instructions:\\
Temporal Logic as a Task Language for VLA Models\\
with Visual-Logical Fusion}

\author{%
  \normalsize Zezhi Liu\textsuperscript{1,2}\quad
  Jingwen Sun\textsuperscript{3,2}\quad
  Hongjin Chen\textsuperscript{4,2}\quad
  Shaojie Ren\textsuperscript{5,2}\\[0.25em]
  \normalsize Peiyan Tu\textsuperscript{6,2}\quad
  Xin Jin\textsuperscript{7,2}\quad
  Zhibo Chen\textsuperscript{3,2}\\[0.9em]
  \small \textsuperscript{1}Nankai University\qquad
  \textsuperscript{2}Beijing Zhongguancun Academy\\
  \small \textsuperscript{3}University of Science and Technology of China\qquad
  \textsuperscript{4}Harbin Institute of Technology\\
  \small \textsuperscript{5}University of the Chinese Academy of Sciences\qquad
  \textsuperscript{6}Zhejiang University\\
  \small \textsuperscript{7}Eastern Institute of Technology, Ningbo
}

\date{}

\hypersetup{
  pdftitle={Beyond Natural-Language Instructions: Temporal Logic as a Task Language for VLA Models with Visual-Logical Fusion},
  pdfauthor={Zezhi Liu, Jingwen Sun, Hongjin Chen, Shaojie Ren, Peiyan Tu, Xin Jin, Zhibo Chen}
}

\begin{document}
\maketitle
\begin{abstract}
Robotic tasks combine object relations, logical dependencies, and temporal requirements, which robot planning and industrial control commonly specify in temporal logic (TL). We introduce \method, a vision--language--action framework that replaces natural-language task instructions with temporal logic, making this spatiotemporal structure explicit. A typed graph encoder and visual--logical fusion condition continuous actions on the task specification and the current observations. To prepare a general vision--language backbone for this interface, we develop TL-aware robot pretraining with separate semantic and structural readouts: a frozen language teacher aligns task semantics, the structural readout encodes TL syntax to support complex, long-horizon tasks, and autoregressive action prediction connects the graph representation to robot behavior. Downstream policies inherit the pretrained backbone and TL encoder and learn new fusion and action modules. On CALVIN ABC$\rightarrow$D, our TL-aware initializer reaches \textbf{3.800} average chain length and \textbf{62.3\%} five-task success, compared with 3.578 and 54.5\% for a FAST initializer pretrained on the same data and 3.485 average chain length for a Qwen+NL baseline. On LIBERO-10, TL-aware initialization improves success from \textbf{91.2\% to 95.8\%}. On LIBERO-Mem, our policy reaches \textbf{65.0\%} success without observation history, compared with 45.0\% for a Qwen+NL reference. On held-out Panda planning specifications, it reaches \textbf{59.6\%} at Success@256 over three generation seeds, versus 48.4\% for Qwen+TL and 37.0\% for Qwen+NL. These results support explicit logical task conditioning and transferable TL-aware representations for robot action learning.
\end{abstract}

\begin{figure*}[t]
\centering
\includegraphics[width=\textwidth]{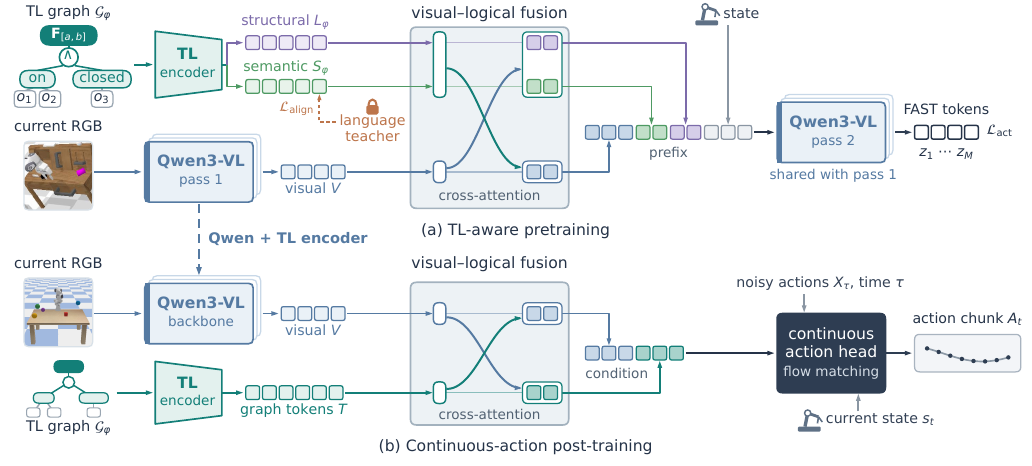}
\caption{\textbf{\method: from structured task intent to continuous actions.} (a) Semantic and structural graph readouts combine with visual features for action-token pretraining; a frozen language teacher (padlock) aligns only the pre-fusion semantic readout. (b) The pretrained Qwen backbone and TL encoder transfer; the fusion module and action head are learned on downstream demonstrations from the current observation, task graph, and robot state. Scene thumbnails are illustrative simulator observations. The schematic depicts the TL policy; NL controls use task text. Solid arrows carry features; dashed arrows mark training-only supervision or weight transfer.}
\label{fig:pipeline}
\end{figure*}

\section{Introduction}
Vision--language--action (VLA) models connect broad visual and linguistic knowledge to robot control. RT-2, OpenVLA, and $\pi_0$ show the value of this connection for learning diverse manipulation behaviors~\cite{rt2,openvla,pi0}. A central design choice is how a task enters the policy. Beyond naming an action, a task may specify which objects participate, how they relate, and which conditions must hold in a particular order. A short natural-language instruction often leaves some of this spatiotemporal structure implicit; we instead make it the explicit task input of the policy.

Consider placing an object in a drawer and then closing the drawer. Temporal logic (TL) represents semantic intent together with spatial relations, temporal dependencies, and compositional syntax. Predicate arguments bind participants; operators specify how requirements relate. This exposes richer, rule-governed structure than a short benchmark instruction, although natural language can express the same requirements. TL is less broadly used than natural language but is well established in robotics, control, and industrial systems. Its precise semantics provide a practical foundation for specifying task and system requirements and supporting systematic planning, control, and monitoring~\cite{lahijanian2012temporal,bttldmps,chen2021temporal,yin2024formal}. This motivates bringing TL into VLA models: \emph{can an explicit logical task interface, supported by suitable pretraining, improve learned visuomotor behavior?}

We introduce \method, which conditions actions on a TL graph, camera observations, and robot state. The graph represents task intent, and perception supplies the current scene. A typed graph encoder preserves operator structure and predicate arguments, and bidirectional visual--logical fusion connects them to visual features. The resulting continuous-action policy takes the graph, rather than an instruction, as its task input; training instructions only supply semantic supervision through a frozen teacher.

Because TL conveys both semantic intent and explicit spatiotemporal syntax, our pretraining splits graph representations into semantic and structural streams. The semantic stream aligns with task-matched training instructions, while the structural stream retains ordered graph information beyond that alignment target. Joint action-token prediction trains a Qwen3-VL backbone~\cite{qwen3vl} to use these representations on a robot-data mixture. We then selectively transfer the Qwen backbone and the TL graph encoder, and learn the visual--logical fusion module and continuous-action head on downstream demonstrations.

We evaluate the capabilities this design targets: sustained execution of consecutive subtasks on CALVIN, long-horizon transfer on LIBERO-10, multi-stage manipulation without observation history on LIBERO-Mem, and logical satisfaction of held-out single-goal, multi-goal, sequential, and partial-order specifications on the Panda benchmark. Under a fixed task interface and downstream recipe, TL-aware initialization improves over original-Qwen and robot-pretrained initializers, and on Panda its gains concentrate on multi-goal, sequential, and partial-order specifications.

Our contributions are threefold:
\begin{enumerate}
\item \textbf{Replacing natural language with temporal logic in VLA policies, with visual--logical fusion.} 
We adopt structured TL specifications as the policy's sole task input and integrate them with visual observations through bidirectional fusion, conditioning continuous action generation on both task logic and scene information.
\item \textbf{TL-aware pretraining for visual--logical policy learning.}
We develop a multimodal pretraining framework that prepares TL-aware representations for visual--logical integration and transfers them to downstream policy learning.
\item \textbf{Evaluation across diverse benchmarks.}
Experiments across diverse robotic benchmarks demonstrate the benefits of our approach for downstream control, with gains on tasks involving complex structured specifications.
\end{enumerate}

\section{Related Work}
\paragraph{Vision--language--action learning.}
RT-2 and OpenVLA transfer visual--language knowledge to robot control~\cite{rt2,openvla}; RoboFlamingo adds history modeling through imitation learning~\cite{roboflamingo}. $\pi_0$ and GR00T N1 generate continuous actions~\cite{pi0,groot}. $\pi_{0.5}$ and RDT2 improve generalization through heterogeneous co-training and scaled UMI data, respectively~\cite{pi05,rdt2}. UniVLA learns task-centric latent actions for cross-embodiment transfer~\cite{univla}, while Moto learns motion tokens from videos~\cite{moto}. VLAct develops transferable representations through continued robot pretraining~\cite{vlact}. We build on StarVLA~\cite{starvla}.

\paragraph{Logic-based policies and planning.}
For nested STL, Yu et al. synthesize continuous-time controllers using temporal-logic trees and control barrier functions~\cite{nestedstl}. SDT conditions offline Decision Transformers on STL robustness~\cite{sdt}, while STL-conditioned diffusion policies generate diverse, rule-compliant trajectories~\cite{stldiffusion}. TeLoGraF learns TL-conditioned trajectories with graph encoding and flow matching~\cite{telograf}. Vision-TL-Action grounds coordinate-free TL graphs in images~\cite{liu2026vision}, while S-MSP combines visual observations and STL through a structured mixture of experts~\cite{smsp}. Logic-VLA adds STL requirements to the NL instruction of a quadcopter-navigation VLA, with a pretrained STL graph encoder, supervised adaptation, and preference optimization~\cite{logicvla}; STeP uses STL for language-based planning, monitoring, and replanning~\cite{step}. We instead replace the instruction with TL as the sole task input and pretrain the VLA backbone and TL encoder jointly on robot data. To our knowledge, no prior work evaluates this interface across manipulation and planning benchmarks against language conditioning.

\paragraph{Task structure and long-horizon evaluation.}
CALVIN evaluates manipulation chains, LIBERO studies knowledge transfer, and LIBERO-Mem tests history-dependent manipulation~\cite{calvin,libero,liberomem}. PDDL describes action models and planning problems~\cite{pddl}, while BDDL specifies activity objects and logical initial/goal conditions~\cite{behavior}. LIBERO uses BDDL task definitions to connect language instructions, scene initialization, and success checks~\cite{libero}. For symbolic model construction, pix2pred selects VLM-proposed predicates to learn world models from demonstrations~\cite{pix2pred}, while PDDL-ART generates PDDL domains and problems from a demonstration, language, and action names~\cite{pddlart}. Connections include finite-trace LTL-to-PDDL compilation~\cite{baier2006planning}, BDDL grounding within an LTL interface~\cite{eai}, and NL-to-STL translation for cooperative robots in Seeing, Saying, Solving~\cite{seeingsayingsolving}. Our TL graphs expose benchmark-defined goals and temporal structure to the policy.

\section{Temporal Logic as the Policy Task Language}
\label{sec:language}
\subsection{Specification and observation interfaces}
At control time $t$, let $o_t=(I_t^{1:V},s_t)$ contain the available camera views and proprioceptive state. A task is given as a bounded temporal-logic formula $\varphi$. The policy predicts an action chunk
\begin{equation}
 A_t = (a_t,\ldots,a_{t+H_a-1})\sim \pi_\theta(\cdot\mid o_t,\mathcal{G}_\varphi),
\end{equation}
where $\mathcal{G}_\varphi$ is a typed syntax graph. The environment executes a prefix of this chunk before the policy receives a new observation. The task graph specifies the desired behavior, and observations provide current evidence.

We use predicates such as $\operatorname{on}(o_i,o_j)$ or $\operatorname{closed}(o_k)$, logical composition, and bounded temporal operators. A representative grammar is
\begin{equation}
\begin{split}
\varphi ::= {}& p(\mathbf{o})\mid \neg\varphi\mid\varphi\wedge\varphi\mid\varphi\vee\varphi\\
 &\mid\F_{[a,b]}\varphi\mid\G_{[a,b]}\varphi\mid\varphi\U_{[a,b]}\varphi.
\end{split}
\end{equation}
On a discrete state trace, $\F_{[a,b]}p$ requires $p$ at some time in the interval, and $\G_{[a,b]}p$ requires it throughout the interval. An ordered eventuality expresses attaining one condition before another. Benchmark adapters define the concrete predicates and intervals of each task.

\begin{figure}[t]
\centering
\includegraphics[width=\columnwidth]{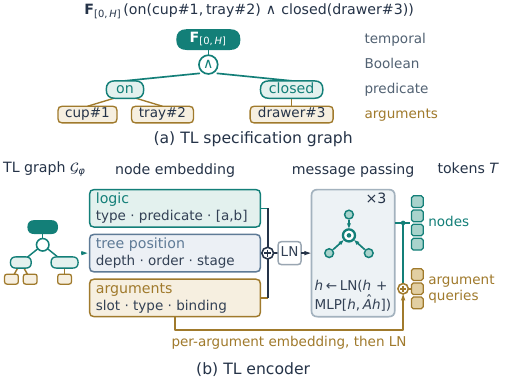}
\caption{\textbf{A TL specification and its encoding.} (a) An illustrative eventual conjunction: both predicates must hold at the same time within the interval; operators and ordered arguments are preserved. (b) Each node sums logic, tree-position, and argument embeddings, then aggregates messages from its tree neighbors for three steps; each argument also yields a query token.}
\label{fig:graph}
\end{figure}

\subsection{Graph encoding and visual binding}
Each graph node carries its logical or predicate type and structural metadata (Figure~\ref{fig:graph}b). The encoder reads normalized temporal intervals, ordered child relations, depth, predicate identity, object type, task-local object slot, occurrence identity, and stage identity. Occurrence identity separates repeated mentions of an object, and stage identity indexes a subformula's position in the specified order; both are fixed formula attributes, not progress signals. 
Tree edges and child order preserve logical branches and asymmetric operator arguments, and object slots identify task-local arguments together with their types and bindings.

Message passing produces contextual node representations; pretraining uses three steps, and the transferred downstream encoder keeps them. Argument-specific query tokens accompany predicate nodes, so the arguments of a spatial relation remain separately addressable. Graph vocabulary and dimensionality are fixed within each initialization comparison. Spatial evidence comes from observations and robot state.

Let $T$ denote graph tokens and $V$ denote selected visual tokens from Qwen. Bidirectional cross-attention (detailed in the appendix) produces fused representations
\begin{equation}
 (\widetilde{T},\widetilde{V})=\operatorname{Fuse}(T,V).
\end{equation}
Graph-to-vision attention supports task-dependent visual selection, and vision-to-graph attention conditions the logical representation on the observed scene. The action head uses these fused features together with proprioception. Where the backbone requires a textual wrapper, we use a fixed, task-independent prompt, so the task condition comes from the graph.

\subsection{What the interface makes explicit}
This formulation separates task intent from any particular language string. Logical edges expose conjunction, alternatives, and temporal structure; predicate arguments expose the participants of each relation. Changing a destination, for example, changes an argument binding rather than the operator structure; Panda tests such recombination on held-out specifications (Sec.~\ref{sec:panda}). In deployment, specifications can come from formal requirements that robotic and industrial systems already maintain~\cite{lahijanian2012temporal,yin2024formal}, planners for temporally extended goals~\cite{baier2006planning}, or NL-to-TL translation~\cite{seeingsayingsolving}. Here, benchmark adapters build them from task definitions and native monitors score success, so task construction, policy input, and evaluation share one specification.

\section{TL-Aware Pretraining and Action Learning}
\label{sec:learning}
\subsection{Pretraining data and multimodal conditioning}
The pretraining backbone is Qwen3-VL-4B-Instruct. The robot mixture samples uniformly from Panda, CALVIN, LIBERO, and LIBERO-Mem. Each adapter supplies current observations, a task graph, native actions, and a source-specific control identity. Every source contributes only training data: CALVIN uses the ABC split and holds out environment D, and Panda excludes the evaluation specifications. 

Images occupy two view slots, and states are padded to 32 dimensions; validity masks mark the available inputs. Actions use source-specific normalization and an eight-step horizon. Embodiment/control and horizon tokens expose these conventions to the autoregressive model.

The graph encoder produces complementary semantic and structural readouts, $S_\varphi$ and $L_\varphi$, to retain task meaning and formal spatiotemporal structure. The latter adds ordered pooling of graph features, preserving the graph sequence independently of semantic alignment. After visual--logical fusion, the autoregressive prefix concatenates the visual features, the two graph-derived streams, state, control identity, and horizon. These features are projected to the Qwen hidden width and receive type embeddings. The action output is encoded with FAST~\cite{fast}, a widely used action tokenization for VLA pretraining~\cite{pi05}. FAST represents action chunks as discrete tokens, so action learning keeps the backbone's next-token prediction objective. To separate the effect of our objectives from exposure to embodied data, we also pretrain PI-Fast-pretrain with FAST action prediction alone. It uses the same data mixture, number of updates, task conditioning, and checkpoint-selection rule as TL-pretrain, and adds no semantic or structural objective.

\subsection{Action prediction and semantic supervision}
For a tokenized action sequence $z_{1:M}$ and multimodal condition $c$, the action objective is
\begin{equation}
\mathcal{L}_{\mathrm{act}}=-\frac{1}{M}\sum_{j=1}^{M}\log p_\theta(z_j\mid z_{<j},c).
\end{equation}
Task language serves only as supervision, through a frozen teacher bank built from training instructions; it never replaces the graph as the policy's task input. The pre-fusion semantic readout is aligned to teacher features, while the structural readout carries an ordered pooled graph residual without a direct language-matching objective.

The alignment objective has token, sentence, and contrastive components:
\begin{equation}
\mathcal{L}_{\mathrm{align}}=\mathcal{L}_{\mathrm{tok}}+0.25\mathcal{L}_{\mathrm{sent}}+0.1\mathcal{L}_{\mathrm{con}}.
\end{equation}
Teacher features are centered with training-bank statistics. Token matching takes the minimum cosine discrepancy over registered instructions of the same task class and length. Sentence matching uses the corresponding semantic-class positives, and a fixed-bank contrastive term separates them from other semantic classes at temperature 0.1. Padding and invalid teacher entries are masked. This set-valued construction admits alternative phrasings of the same task class.

The combined objective is
\begin{equation}
\begin{aligned}
\mathcal{L}_{\mathrm{pre}}&=\mathcal{L}_{\mathrm{act}}+\lambda(u)\mathcal{L}_{\mathrm{align}},\\
\lambda(u)&=0.1\min\!\left(1,\frac{u+1}{1000}\right),
\end{aligned}
\end{equation}
where $u$ is the optimizer-update index. During pretraining, the graph readout length follows the paired instruction-token count, capped at 64 tokens, which fixes the alignment layout. Downstream policies take graph tokens directly from the transferred encoder, without the readouts, teacher, or instruction-token count.

Pretraining uses global batch 256, seed 42, AdamW with $(\beta_1,\beta_2)=(0.9,0.95)$, and a cosine schedule with 2,000 warmup updates. The Qwen peak learning rate is $10^{-6}$; the newly learned graph and conditioning modules use $10^{-4}$. We save initializers along this single 50,000-update run.

\subsection{Selective transfer and continuous actions}
We selectively transfer the pretrained Qwen backbone and TL graph encoder, and learn the visual--logical fusion module and action head on downstream demonstrations. Pretrained representations are reused, while cross-modal interaction adapts to downstream observations and control. Token embeddings are restored to the original backbone vocabulary, and the encoder keeps its pretrained configuration. The readouts, teacher, pretraining fusion, and FAST decoder are not transferred.

We use a GR00T-style diffusion-transformer action head trained with conditional flow matching~\cite{groot,flowmatching}. For a normalized demonstrated action chunk $A$ and Gaussian noise $\epsilon$, let
\begin{equation}
X_\tau=(1-\tau)\epsilon+\tau A,\qquad u=A-\epsilon.
\end{equation}
The policy learns a conditional velocity field with a masked squared error,
\begin{equation}
\mathcal{L}_{\mathrm{down}}=\E\!\left[\frac{\|m\odot(v_\theta(X_\tau,\tau,o_t,\mathcal{G}_\varphi)-u)\|_2^2}{\|m\|_1}\right],
\end{equation}
where the masked residual is concatenated over a normalization group and its repeated noise draws, the outer average is over groups, and $m$ excludes invalid action elements. With $b\sim\operatorname{Beta}(1.5,1)$, we sample $\tau=(s-\min(b,s))/s$ with $s=0.999$. The matched CALVIN and LIBERO controls use groups of four and eight observations, respectively, eight training noise draws per observation, and four inference steps. Panda and LIBERO-Mem use two.

The downstream head has width 1,024, 16 transformer layers, and seven-dimensional actions. The Qwen feature width is 2,560, and TL fusion uses eight attention heads. CALVIN and LIBERO both train for 30,000 updates with 5,000 warmup updates. The Qwen and action/TL peak learning rates are $10^{-5}$ and $10^{-4}$, respectively, with a cosine schedule to a $10^{-6}$ floor. The backbone and TL encoder remain trainable. Benchmark-specific schedules are in the appendix.

\section{Experiments}
\label{sec:experiments}
\begin{figure*}[t]
\centering
\includegraphics[width=\textwidth]{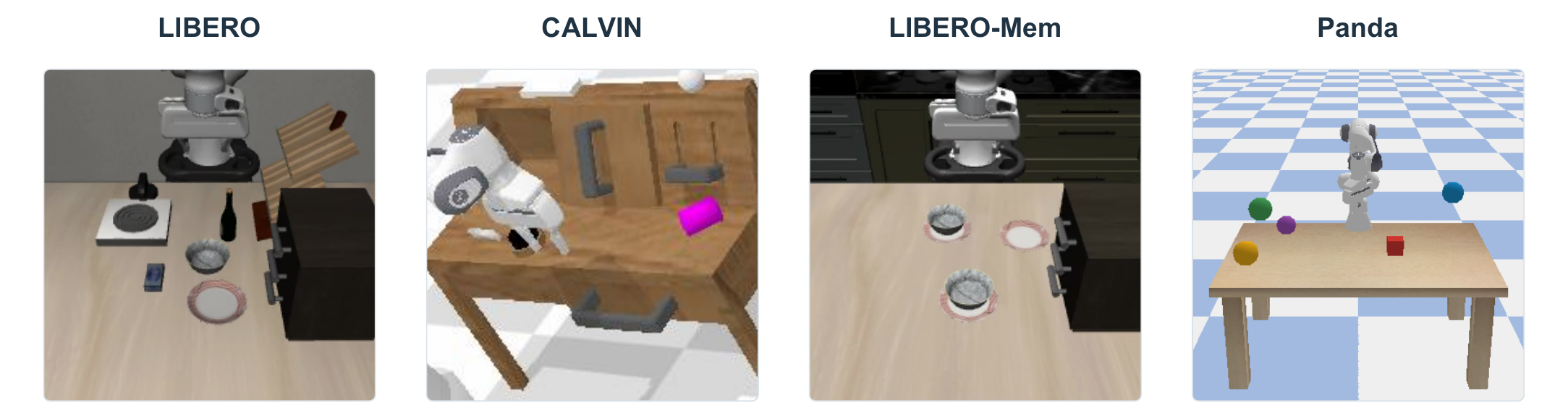}
\caption{\textbf{Four post-training environments and benchmarks.} Representative simulator observations from LIBERO, CALVIN, LIBERO-Mem, and Panda.}
\label{fig:environments}
\end{figure*}

We evaluate four downstream capabilities: sustained execution of consecutive subtasks on CALVIN, long-horizon task execution on LIBERO-10, sequential manipulation without observation history on LIBERO-Mem, and logical satisfaction of held-out specifications on Panda (Figure~\ref{fig:environments}).

\paragraph{Task information and comparison scope.}
Benchmark adapters build each TL graph automatically from the task's NL description and native symbolic metadata (BDDL where available; examples in the appendix). The graph is fixed per episode: it states goal conditions, bindings, and order, not which conditions already hold, so the policy infers progress from observations. NL instructions come from the same task definitions but may leave bindings and order implicit; making them explicit is the property under test, so TL-versus-NL rows compare complete pipelines. The graph alone is no oracle: with original Qwen, TL trails NL on CALVIN (Table~\ref{tab:calvin}). Rows sharing identical graphs isolate TL-aware pretraining. CALVIN and LIBERO evaluate training tasks in new environments or initial states and were not designed to vary formula structure; Panda tests held-out formulas  \cite{telograf}. 95\% intervals for key differences use evaluation counts (normal approximation) and cover evaluation sampling only; each policy is trained once (seed 42).

\subsection{CALVIN: main long-horizon comparison}
\label{sec:calvin}
\paragraph{Comparison design of task and evaluation.}
CALVIN ABC$\rightarrow$D tests whether a policy can sustain manipulation across five consecutive subtasks in an unseen environment~\cite{calvin}. The official evaluator issues one subtask at a time and advances once its oracle reports success, so NL rows receive the current instruction and TL rows the current subtask's graph. CALVIN thus measures sustained execution from the states earlier subtasks leave, not parsing of a five-task formula; LIBERO-Mem and Panda supply complete specifications. 
We compare Qwen+NL, S-MSP~\cite{smsp}, Logic-VLA~\cite{logicvla}, Qwen+TL, PI-Fast-pretrain+NL, PI-Fast-pretrain+TL, and TL-pretrain+TL (Table~\ref{tab:calvin}). The three TL-conditioned initialization controls share the task interface, action head, and downstream recipe: 30k updates, global batch 32, and training seed 42. They differ only in initialization: original Qwen or PI-Fast-pretrain, each with a new TL encoder, or TL-pretrain, which supplies both Qwen and the TL encoder. The NL and external rows are whole-method comparisons; we adapt S-MSP and Logic-VLA to CALVIN, keeping Logic-VLA's instruction and STL inputs (appendix). 
The table is a controlled study of task interfaces and initializers, not a leaderboard; methods that redesign the visual and action pathways reach 4.28--4.53 ACL~\cite{seer,vpp,flower}, and those designs are orthogonal to the task interface studied here.

All TL controls use the same 1,000 official sequences, evaluation seed 0, and at most 360 simulator steps per subtask. At each replanning step, the policy predicts eight actions and executes five before receiving a new observation. For completed chain length $c_i\in\{0,\ldots,5\}$, we report average completed chain length (ACL) and success on at least $k$ consecutive tasks:
\begin{equation}
\operatorname{ACL}=\frac{1}{N}\sum_i c_i,\quad
S_k=\frac{100}{N}\sum_i\mathbb{1}[c_i\geq k].
\end{equation}
Consequently, $\operatorname{ACL}=\sum_{k=1}^{5}S_k/100$.

\begin{table*}[t]
\centering\small
\setlength{\tabcolsep}{7pt}
\caption{\textbf{CALVIN ABC$\rightarrow$D: main long-horizon comparison.} $S_k$ is success (\%) on at least $k$ consecutive tasks; ACL is average completed chain length. Latency is the mean time per policy call. Model sources, planning horizons, and timing boundaries are detailed in the appendix.}
\label{tab:calvin}
\begin{tabular*}{\textwidth}{@{\extracolsep{\fill}}lrrrrrrr@{}}
\toprule
Method & ACL $\uparrow$ & $S_1$ & $S_2$ & $S_3$ & $S_4$ & $S_5$ & \shortstack{Latency $\downarrow$\\(ms/call)} \\
\midrule
Qwen + NL & 3.485 & 91.8 & 80.3 & 69.6 & 58.2 & 48.6 & 289.5 \\
PI-Fast-pretrain + NL & 3.610 & 90.6 & 82.4 & 71.3 & 63.2 & 53.5 & 288.9 \\
S-MSP \cite{smsp} & 1.755 & 66.4 & 46.3 & 30.6 &19.5 & 12.7 & 420.9 \\ 
Logic-VLA \cite{logicvla} & 3.686 & \textbf{92.7} & \textbf{84.2} & 73.4 & 64.1 & 54.2 & 223.5 \\
\midrule
Qwen + TL & 2.685 & 80.4 & 64.5 & 51.7 & 41.1 & 30.8 & 103.3 \\
PI-Fast-pretrain + TL & 3.578 & 90.7 & 80.2 & 70.3 & 62.1 & 54.5 & 100.0 \\
\textbf{TL-pretrain + TL (ours)} & \textbf{3.800} & 89.9 & 82.4 & \textbf{76.1} & \textbf{69.3} & \textbf{62.3} & \textbf{99.4} \\
\bottomrule
\end{tabular*}
\end{table*}

\begin{figure*}[t]
\centering
\includegraphics[width=\textwidth]{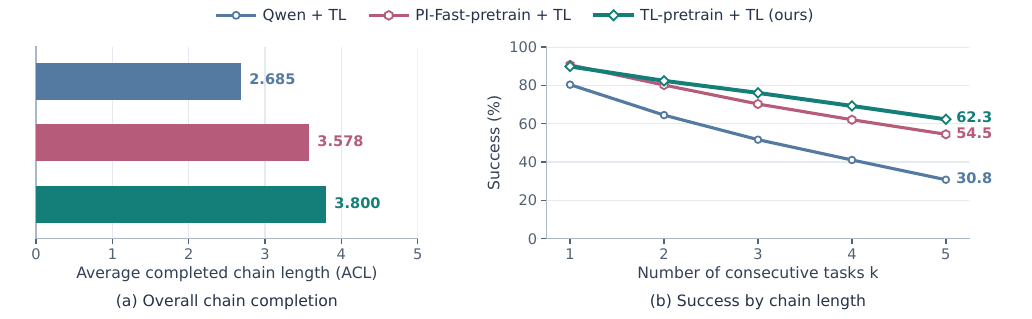}
\caption{\textbf{CALVIN initialization controls under the same TL interface.} The curves share the downstream recipe and the 1,000 official sequences and differ only in initialization. TL-pretrain leads at chain lengths two through five, with a growing advantage over PI-Fast-pretrain, which uses the same robot data.}
\label{fig:calvin}
\end{figure*}

\paragraph{Main result: stronger completion of long chains.}
TL-pretrain+TL achieves the highest ACL (3.800) and the highest three-, four-, and five-task success in Table~\ref{tab:calvin}. Its five-task success of 62.3\% exceeds Logic-VLA by 8.1 percentage points, with an ACL gain of 0.114. Relative to Qwen+NL, the complete approach improves ACL by 0.315 and five-task success by 13.7 points.

Under the shared TL interface, TL-aware initialization improves ACL from 2.685 to 3.800 and five-task success from 30.8\% to 62.3\% over original Qwen. Against PI-Fast-pretrain, which uses the same robot data, TL-pretrain gains 0.222 ACL (95\% CI 0.06--0.38) and 7.8 points of five-task success (CI 3.5--12.1), or 78 additional complete chains. The gap grows from 2.2 points at two tasks to 7.8 at five because a per-step margin compounds: after a completed subtask, TL-pretrain completes the next in 89.9--92.4\% of cases, versus 87.7--88.4\% for PI-Fast-pretrain (Figure~\ref{fig:calvin}). Since subtasks arrive one at a time, this is evidence of more reliable continued execution, and it supports preparing the backbone and graph encoder for TL-conditioned control.

The CALVIN TL policies read fused visual and graph features directly from the backbone and take 99.4--103.3\,ms per call, versus 289.5\,ms for Qwen+NL ($2.8$--$2.9\times$ faster; details in the appendix).

\subsection{LIBERO: long-horizon transfer and robustness}
\label{sec:libero}
TL-aware pretraining improves long-horizon execution on LIBERO-10~\cite{libero} (Table~\ref{tab:libero}). TL-pretrain+TL reaches 95.8\% success, 4.6 percentage points above Qwen+TL (91.2\%; 95\% CI of the gain 1.6--7.6) and above all StarVLA baselines (88.4--93.8\%). The matched TL pair shares the task graphs, current observations, and 30k-update downstream recipe; only initialization changes.

\begin{table*}[!t]
\centering
\begin{minipage}[t]{0.48\textwidth}
\vspace{0pt}
\centering\small
\renewcommand{\arraystretch}{1.12}
\setlength{\tabcolsep}{0pt}
\caption{\textbf{LIBERO-10: long-horizon manipulation.} Matched TL pair: 500 trials each. Protocols are given in the appendix.}
\label{tab:libero}
\begin{tabular}{@{}>{\raggedright\arraybackslash}p{0.49\linewidth}>{\raggedleft\arraybackslash}p{0.24\linewidth}>{\raggedleft\arraybackslash}p{0.27\linewidth}@{}}
\toprule
Method & Successes & Success (\%) $\uparrow$ \\
\midrule
StarVLA Qwen3 + FAST \cite{starvla} & 453 / 500 & 90.6 \\
StarVLA Qwen3 + OFT \cite{starvla} & 469 / 500 & 93.8 \\
StarVLA Qwen3 + PI \cite{starvla} & 442 / 500 & 88.4 \\
StarVLA Qwen3 + GR00T \cite{starvla} & 460 / 500  & 92.0 \\
Qwen + TL & 456 / 500 & 91.2 \\
\textbf{TL-pretrain + TL (ours)} & \textbf{479 / 500} & \textbf{95.8} \\
\bottomrule
\end{tabular}
\end{minipage}\hfill
\begin{minipage}[t]{0.48\textwidth}
\vspace{0pt}
\centering\small
\renewcommand{\arraystretch}{1.12}
\setlength{\tabcolsep}{0pt}
\caption{\textbf{LIBERO-Plus-10: robustness.} Canonical TL graphs; 2,519 variants per policy. Sem.: separate variant with a semantic middle stage (Sec.~\ref{sec:libero}).}
\label{tab:libero_plus}
\begin{tabular}{@{}>{\raggedright\arraybackslash}p{0.49\linewidth}>{\raggedleft\arraybackslash}p{0.24\linewidth}>{\raggedleft\arraybackslash}p{0.27\linewidth}@{}}
\toprule
Method & Successes & Success (\%) $\uparrow$ \\
\midrule
Qwen2.5 + NL(GR00T) \cite{starvla} & 1471 / 2519 & 58.40  \\
Qwen3 + TL & 1671 / 2519 & 66.34 \\
TL-pretrain + TL & 1744 / 2519 & 69.23 \\
TL-pretrain + TL + Sem. & 1819 / 2519 & \textbf{72.21} \\
\bottomrule
\end{tabular}
\end{minipage}
\end{table*}

A separate variant, TL-pretrain+TL+Sem., adds a semantic middle stage (shown in appendix); it lies outside the matched comparisons and is not used elsewhere. Across Spatial, Object, Goal, and LIBERO-10, it reaches 95.75\% (1,915/2,000), comparable to NL-conditioned policies. As LIBERO tasks carry limited logical structure, we read this aggregate as broad manipulation competence.

LIBERO-Plus-10~\cite{liberoplus} tests robustness to perturbed observations and initial conditions with fixed canonical TL graphs (Table~\ref{tab:libero_plus}). TL pretraining raises success from 66.34\% (Qwen + TL) to 69.23\% (95\% CI of the gain 0.3--5.5), and the Sem.\ variant reaches 72.21\%. Per-task results are in the appendix.

\subsection{LIBERO-Mem: sequential manipulation}
\label{sec:libero_mem}
On LIBERO-Mem~\cite{liberomem}, we test whether the TL-pretrained policy executes multi-stage goals without observation history. TL-pretrain+TL receives the complete, fixed task graph, dual-view RGB, and joint state, without observation history or online memory. The graph does not mark completed subgoals, so the policy infers the active stage from the current views. We evaluate after 20k downstream updates on ten tasks with 20 trials each and, matching our design, compare only with memory-free methods.

\begin{table*}[!t]
\centering\small
\renewcommand{\arraystretch}{1.15}
\setlength{\tabcolsep}{3pt}
\caption{\textbf{LIBERO-Mem (\%).} Per-task success rates of memory-free policies. TL-pretrain + TL achieves the highest average success (65.00\%), 20 percentage points above Qwen + NL, with the largest gains on \textbf{Tasks 6 and 7}, the two ordered rearrangements.}
\label{tab:libero_mem_main}
\begin{tabular*}{\textwidth}{@{\extracolsep{\fill}}lrrrrrrrrrrr@{}}
\toprule
Method & Task 0 & Task 1 & Task 2 & Task 3 & Task 4 & Task 5 & \textbf{Task 6} & \textbf{Task 7} & Task 8 & Task 9 & Avg. $\uparrow$ \\
\midrule
$\pi_{0.5}$ (STVLA)~\cite{stvla} & 95.00 & 5.00 & 80.00 & 0.00 & 40.00 & 10.00 & 10.00 & 5.00 & \textbf{55.00} & \textbf{80.00} & 38.00 \\
Qwen + flow (relaxed)~\cite{twoam} & 71.67 & \textbf{91.67} & 52.50 & 40.00 & 42.50 & 54.17 & 6.67 & 9.17 & 0.00 & 5.83 & 37.42 \\
Qwen + NL & 70.00 & 90.00 & 60.00 & \textbf{55.00} & 45.00 & 60.00 & 5.00 & 0.00 & 25.00 & 40.00 & 45.00 \\
\midrule
\textbf{TL-pretrain + TL (ours)} & \textbf{100.00} & 90.00 & \textbf{85.00} & 40.00 & \textbf{50.00} & \textbf{65.00} & \textbf{60.00} & \textbf{55.00} & 50.00 & 55.00 & \textbf{65.00} \\
\bottomrule
\end{tabular*}
\end{table*}

TL-pretrain+TL completes 130/200 trials (65.0\%), 20.0 percentage points above the Qwen+NL mean of 45.0\% (95\% CI of the gain 10.4--29.6; Table~\ref{tab:libero_mem_main}). Tasks 6 and 7 are the two lowest-scoring tasks for the NL reference: success rises from 5\% to 60\% and from 0\% to 55\%, a gain of 55 points on each. The published memory-free methods, whose results we take from their papers, also score low on these two rearrangements.

Task 6 swaps two bowls through an empty plate via either of two three-stage routes. Task 7 rotates three bowls from left to right through the empty plate in a fixed four-stage sequence. The temporary placement frees a destination for later transfers. These gains are consistent with explicit object bindings and ordered subgoals guiding the transfers, while the current RGB grounds action selection and stage tracking. Where NL leaves this order implicit, the gain credits the interface: a stated order plus current views replaces part of the required memory.

\begin{figure*}[!t]
\centering
\includegraphics[width=\textwidth]{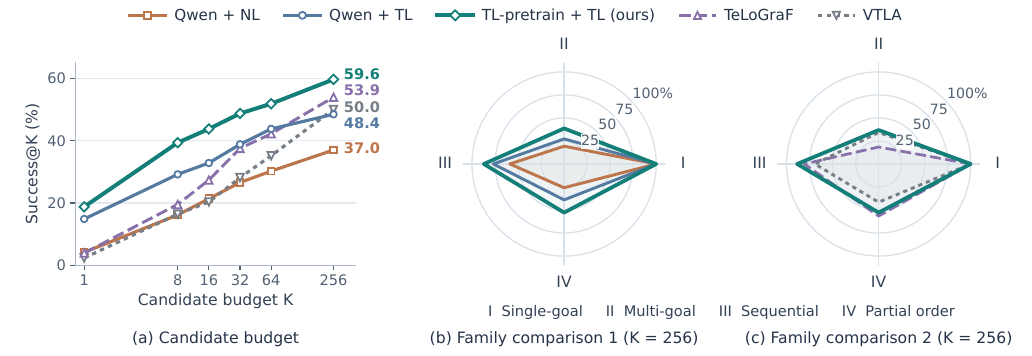}
\caption{\textbf{Panda: candidate budgets and logical task families.} (a) Offline-oracle Success@$K$ and (b) family-wise Success@256 for the three internal models, aggregated over three generation seeds. (c) Success@256 for TL-pretrain + TL (ours), TeLoGraF~\cite{telograf}, and VTLA~\cite{liu2026vision} on the same 128 tasks. Both radars normalize by family size.}
\label{fig:panda}
\end{figure*}

\subsection{Panda: spatiotemporal-logic satisfaction}
\label{sec:panda}
\paragraph{A logic-centered benchmark.}
We adapt the Franka Panda benchmark from TeLoGraF~\cite{telograf} to visual task conditioning. Its four specification families, \emph{I. single-goal}, \emph{II. multi-goal}, \emph{III. sequential}, and \emph{IV. partial order}, cover timed goal reaching, Boolean goal combinations, ordered visits, and partial-order dependencies, all with obstacle-avoidance constraints. Beyond $\F$ and $\wedge$, the templates use $\G\neg$ for avoidance, $\vee$ for alternative goals, and $\neg\,\U$ for precedence~\cite{telograf}. The 128 evaluation specifications are held out from training, so they require new combinations of operators, bindings, and intervals. Each policy gets the whole formula and orders subgoals itself under partial orders.

We compare Qwen+NL, Qwen+TL, and TL-pretrain+TL on all 128 tasks with three generation seeds. For these three models, dual-view images are re-rendered from each candidate's current generated joint state every eight actions. Success@$K$ is an \emph{offline oracle}: a task--seed pair succeeds if any of its first $K$ candidates satisfies the native STL evaluator. The three models share the candidate-noise bank and a 21,399-update training budget, and both TL models share the downstream recipe. The appendix details the NL comparison and the archived planner references.

\begin{table*}[t]
\centering\small
\setlength{\tabcolsep}{2.5pt}
\caption{\textbf{Panda Success@$K$ (\%)} under different candidate budgets. TL-pretrain + TL achieves the highest success at every $K$.}
\label{tab:panda}
\begin{tabular*}{\textwidth}{@{\extracolsep{\fill}}lrrrrrr@{}}
\toprule
Method & @1 & @8 & @16 & @32 & @64 & @256 \\
\midrule
TeLoGraF \cite{telograf} & 3.91 & 19.53 & 27.34 & 37.50 & 42.19 & 53.91 \\
VTLA \cite{liu2026vision}  & 2.34 & 16.41 & 20.31 & 28.13 & 35.16 & 50.00 \\
\midrule
Qwen + NL & 4.17 & 16.15 & 21.35 & 26.56 & 30.21 & 36.98 \\
Qwen + TL & 14.84 & 29.17 & 32.81 & 38.80 & 43.75 & 48.44 \\
TL-pretrain + TL & \textbf{18.75} & \textbf{39.32} & \textbf{43.75} & \textbf{48.70} & \textbf{51.82} & \textbf{59.64} \\
\bottomrule
\end{tabular*}
\end{table*}

\paragraph{Pretraining and logical task structure.}
TL-pretrain+TL reaches 59.64\% Success@256, exceeding Qwen+TL by 11.20 percentage points (95\% CI 4.2--18.2) and Qwen+NL by 22.66 points (Table~\ref{tab:panda}), and it leads at all six budgets.

Figure~\ref{fig:panda}(b) locates the gains in composed tasks, since single-goal success is 100\% for all three models. Sequential success reaches 87.18\%, versus 76.92\% for Qwen+TL and 58.97\% for NL. TL-pretrain+TL also improves over NL on multi-goal tasks (38.60\% versus 19.30\%) and partial-order tasks (52.83\% versus 25.79\%). This pattern is consistent with explicit bindings, operators, and precedence constraints helping the policy satisfy composed goals.

The planner comparison in Figure~\ref{fig:panda}(c) uses generation seed 2027 for all methods. TL-pretrain+TL achieves 59.38\% overall, versus 53.91\% for TeLoGraF and 50.00\% for VTLA, with sequential success of 88.46\%, 80.77\%, and 69.23\%, respectively. These archived planners use their original open-loop observation and planning protocols, so they provide context rather than a matched comparison.

\section{Discussion and Conclusion}

\paragraph{Discussion.}
The gains of \method{} grow with task length and compositional structure. On CALVIN, where subtasks arrive one at a time, TL-pretrain completes each later subtask more reliably, so its margin over PI-Fast-pretrain widens from 2.2 points at two tasks to 7.8 at five. With complete specifications, gains concentrate on composed Panda families and on the two ordered LIBERO-Mem rearrangements. Structure alone does not suffice: with original Qwen, TL trails NL on CALVIN (2.685 vs.\ 3.485 ACL), while TL-aware initialization lifts it to 3.800. Our evidence is limited to simulation, one training seed, and adapter-built graphs; formula-level generalization is tested on Panda only. Real-robot evaluation, NL-to-TL translation, and integration with symbolic planners~\cite{pddl} are natural next steps.

\paragraph{Conclusion.}
\method{} replaces natural-language instructions with TL as the policy's task input, grounds the task graph in observations through visual--logical fusion, and prepares the model with semantic--structural pretraining. The results suggest that explicit task structure improves long-horizon, logically composed manipulation once the model is pretrained to use it.

\clearpage
{\small
\bibliographystyle{ieeenat_fullname_unsrt}
\bibliography{main}
}
\end{document}